\documentclass[conference]{IEEEtran}
\IEEEoverridecommandlockouts

\usepackage{url}
\usepackage{cite}
\usepackage{amsmath,amssymb,amsfonts}
\usepackage{accents}
\usepackage[linesnumbered,ruled]{algorithm2e}
\usepackage[noend]{algpseudocode}
\usepackage{graphicx}
\usepackage{textcomp}
\usepackage[table,xcdraw]{xcolor}
\usepackage{xcolor}
\usepackage{comment}
\usepackage{multicol,multirow}
\usepackage[T1]{fontenc}
\def\BibTeX{{\rm B\kern-.05em{\sc i\kern-.025em b}\kern-.08em
    T\kern-.1667em\lower.7ex\hbox{E}\kern-.125emX}}
\usepackage{todonotes}
\usepackage{caption}
\usepackage{wrapfig}
\usepackage{subfig}
\newsavebox{\measurebox}
\usepackage{threeparttable}
\usepackage[font=small]{caption}
\usepackage{enumitem}

\usepackage{hyperref}
\hypersetup{
    colorlinks=false
}

\makeatletter
\newcommand{\algorithmfootnote}[2][\footnotesize]{%
  \let\old@algocf@finish\@algocf@finish
  \def\@algocf@finish{\old@algocf@finish
    \leavevmode\rlap{\begin{minipage}{\linewidth}
    #1#2
    \end{minipage}}%
  }%
}
\makeatother

\newcommand*{\Scale}[2][4]{\scalebox{#1}{$#2$}}%

\begin{document}
\bstctlcite{IEEEexample:BSTcontrol}

\title{Data-driven Design of Context-aware Monitors for\\ Hazard Prediction in Artificial Pancreas Systems} 

\author{Regular Paper}

\author{Xugui Zhou, Bulbul Ahmed\IEEEauthorrefmark{1}, James H. Aylor, Philip Asare\IEEEauthorrefmark{2}, Homa Alemzadeh\\

 \{xz6cz,jha,ha4d@virginia.edu\}University of Virginia, Charlottesville, VA 22904, USA \\
 \IEEEauthorrefmark{1}University of Florida, Gainesville, FL 32611, USA
 \IEEEauthorrefmark{2}University of Toronto, Toronto, ON, Canada }

\maketitle
\thispagestyle{plain}
\pagestyle{plain}

\vspace{-2em}
\begin{abstract} 

Medical Cyber-physical Systems (MCPS) are vulnerable to accidental or malicious faults that can target their controllers and cause safety hazards and harm to patients. This paper proposes a combined model and data-driven approach for designing context-aware monitors that can detect early signs of hazards and mitigate them in MCPS. We present a framework for formal specification of unsafe system context using Signal Temporal Logic (STL) combined with an optimization method for patient-specific refinement of STL formulas based on real or simulated faulty data from the closed-loop system for the generation of monitor logic. We evaluate our approach in simulation using two state-of-the-art closed-loop Artificial Pancreas Systems (APS). The results show the context-aware monitor achieves up to 1.4 times increase in average hazard prediction accuracy (F1-score) over several baseline monitors, reduces false-positive and false-negative rates, and enables hazard mitigation with a 54\% success rate while decreasing the average risk for patients.


\end{abstract}


\section{Introduction}
\vspace{-0.5em}
Medical Cyber-Physical Systems (MCPS) are increasingly deployed in various safety-critical diagnostic and therapeutic applications. 
Recent studies have shown the susceptibility of medical devices, such as patient monitors, infusion pumps, implantable pacemakers, and surgical robots to accidental faults or malicious attacks with potential adverse impacts on patients \cite{alemzadeh2013analysis, halperin2008pacemakers,li2011hijacking,Bonacidos, dosattack, alemzadeh2016targeted,alemzadeh2012towards}. Although leveraging correct-by-construction techniques like formal methods, model-based design, and automated synthesis can improve the resilience of CPS, they are still vulnerable to residual faults and attacks that can evade even the most rigorous design and verification methods and appear during run time. 

Run-time verification of safety properties based on formal models of systems has been an active area of research in safety-critical systems \cite{lee1999runtime, eakman2018correct, runtimeFalcone, deshmukh2017robust}. However, these approaches often rely on ad-hoc safety properties and do not account for cyber-physical system interactions and the multi-dimensional context in the CPS. 
Recent works on anomaly detection in CPS rely on complex dynamic models of physical system/environment~\cite{Althoffonline}\cite{alemzadeh2016targeted} and/or human operator actions~\cite{yasar_ismr2019,yasar_dsn2020} for improved detection accuracy and latency~\cite{lin2020challenges}. But developing such dynamic models for MCPS is challenging because of the variety of patient profiles and unpredictable changes in the human body over time. 

Great efforts have also been made to improve the MCPS safety and security using online monitoring and anomaly detection, including model-based approaches \cite{young2018damon, enforcePini}, probabilistic models \cite{rao2017probabilistic}, fuzzy logic-based algorithms \cite{oliveira2015fuzzy}, invariant detection techniques~\cite{aliabadi_artinali_2017, yasar_ismr2019}, and machine learning~\cite{yasar_dsn2020}. 
However, most of these solutions do not provide the ability for \emph{early detection} of safety property violations, which would help with the \emph{prevention} of hazards.

In this paper, we propose a methodology for designing context-aware safety monitors that can detect early signs of safety hazards in MCPS by identifying potentially unsafe cyber-physical interactions. 
Our method combines the formal specification of safety context for run-time monitoring of the MCPS controller's actions with the data-driven optimization of the monitor's logic based on real or simulated patient data to predict impending hazards. What differentiates our method from previous context-aware monitoring solutions~\cite{yasar_ismr2019,yasar_dsn2020,mcps2018} is combining domain knowledge with data to improve detection accuracy, timeliness, and transparency. Our proposed monitor can be integrated with the control software of a target MCPS and only requires access to its input-output interface (sensor and actuator values). We demonstrate the effectiveness of our approach with a case study of Artificial Pancreas Systems (APS) used for diabetes management. 


The main contributions of the paper are as follows:
\begin{itemize}[leftmargin=*]
\item Proposing a framework for formal specification of safety context for hazard prediction and mitigation in MCPS (Section \ref{sec:monitor-synthesis}). This framework closes the gap between design-time hazard analysis and run-time safety monitoring and enables the generation of template signal temporal logic (STL) formulas for run-time identification of unsafe control actions that potentially lead to hazards.
\item Developing a data-driven method for patient-specific refinement of the STL formulas and their translation into monitor logic based on real or simulated faulty data collected from the closed-loop MCPS (Section  \ref{subsec:STL_learning}). Our method uses the L-BFGS-B {\cite{BFGS_Morales}} optimization algorithm with a tight exponential loss function for learning patient-specific parameters in the monitor logic. It shows improved tightness and convergence rate compared to a previous STL learning method and achieves better prediction accuracy compared to traditional machine learning techniques.

\item Developing an open-source environment for experimental evaluation of different monitors in terms of timely and accurate prediction of hazards for the case study of APS (Fig. \ref{fig:FI-framework}a). This environment integrates two different APS controllers, OpenAPS \cite{openSourceOpenAPS} and Basal-Bolus \cite{BBcontroller}, with widely-used patient glucose simulators, Glucosym \cite{openSourceGlucosym} and UVA-Padova \cite{man2014uva}, for closed-loop simulation of APS with 20 diabetic patient profiles (Section \ref{subsec:closed_loop}), as well as a software fault injection (FI) engine for simulation of representative fault and attack scenarios (Section \ref{subsec:threat_model}). 

\item Introducing new metrics for evaluation of real-time anomaly detection techniques in MCPS, including hazard prediction accuracy with a tolerance window, reaction time, recovery rate, and average risk (Section \ref{subsec:metrics}). These metrics measure the impact of detection accuracy and latency on the successful hazard mitigation and prevention of harm to patients.
\item Evaluating the proposed context-aware monitor using two different closed-loop APS systems and simulators, OpenAPS with Glucosym and Basal-Bolus with UVA-Padova, in comparison to several baseline monitors developed using medical guidelines, model predictive control (MPC), and machine learning (ML). Our results (Section \ref{subsec:results}) show that the patient-specific safety monitor developed with this approach demonstrates up to 1.4 times increase in average prediction accuracy  (F1  score)  over baseline monitors, reduces false-positive and false-negative rates, and enables hazard mitigation with a 54\% success rate while decreasing the average risk for patients.
\end{itemize}

\vspace{-0.2em}
\section{Preliminaries}
\begin{figure}[t!]
    \centering
    \includegraphics[trim={0 1cm 0cm 0cm}, width=\columnwidth]{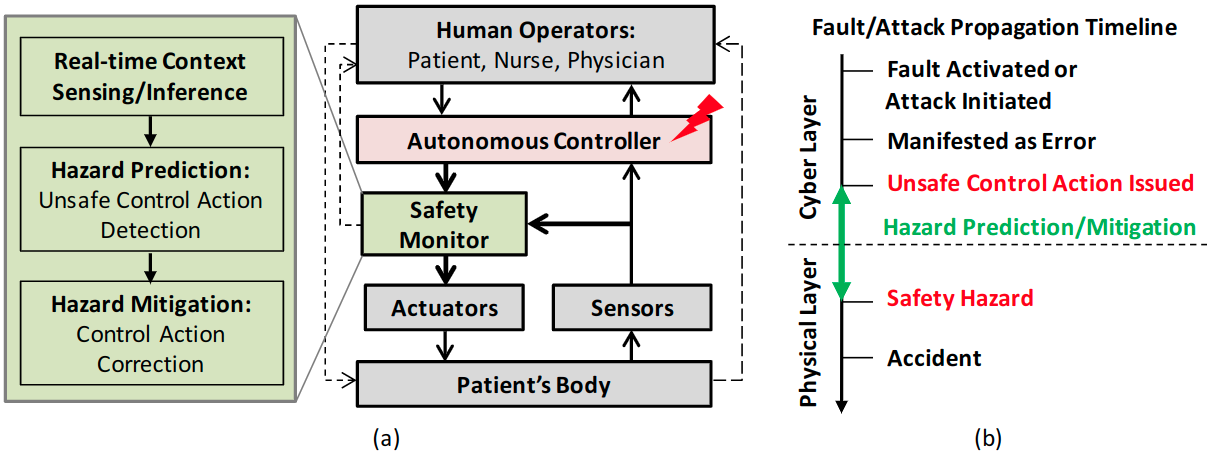}
    \caption{(a) MCPS Control System with the Context-aware Safety Monitor, (b) Fault Propagation Timeline. 
    }
    \label{fig:Figure1}
    \vspace{-1.5em}
\end{figure}

CPS are constructed by the tight integration of cyber components and software with hardware devices and the physical world. The core of the MCPS are the autonomous controllers that connect the human operators (e.g., physicians, nurses) and cyber networks with the physical components (e.g., patient's body) (Fig. \ref{fig:Figure1}a). The controller's goal is to adapt to the constantly changing and uncertain physical environment and the operator's commands by estimating the system's current state based on sensor measurements and changing the physical state by sending control commands to the actuators. In such systems, safety hazards and accidents might happen due to unsafe commands issued by the controller because of accidental faults or malicious attacks acting on the sensor data, the controller (algorithm, software, hardware), or the actuators. 

\textbf{Vulnerable Controllers:} Past studies have emphasized the risks of security attacks that compromise the communication channels in medical devices \cite{halperin2008pacemakers, li2011hijacking, dosattack}.  Safety-critical faults or attacks on sensor data, before they are delivered to the controller, can be detected by previously proposed strategies like redundancy\cite{Mehmed5794}, classic Sequential Probability Ratio Test (SPRT) of Wald\cite{sqrt}, change detection techniques (e.g., Cumulative Sum Control Chart (CUSUM)\cite{attack_pcs}), or well-trained ML models \cite{sadhu2020onboard}\cite{NetworkAnomalyDetection}\cite{Taisa2020}. However, if the attacks do not exhibit malicious behaviors until the controller has received the sensor data, or accidental and malicious faults directly compromise the controller software or hardware functionality, the techniques mentioned above will fail to detect them. This is probable given the existing vulnerabilities in the communication channels of devices \cite{alemzadeh2016targeted}\cite{ramkissoon2017review} and recent trends towards open-source \cite{opensourcechallenge} and mobile and app-based  \cite{apsapp}\cite{FDA_medApps} controllers. In this paper, we aim to address this problem by focusing on the faults/attacks targeting the controller itself. 

\textbf{Hazard Prediction:} Our goal is to detect potentially unsafe control commands issued by an MCPS controller, regardless of their originating causes, and stop or mitigate them \textit{before} execution on the actuators to prevent safety hazards. This is based on the observation that there is a time gap from the activation of the fault and generation of erroneous control commands in the cyber layer until the occurrence of a hazardous state in the physical layer \cite{alemzadeh2016targeted} leading to an accident (Fig. \ref{fig:Figure1}b). As shown in Fig. \ref{fig:Figure1}a, we propose to integrate a safety monitor with a target MCPS controller as a wrapper that only has access to the input-output interface for observing the sensor data and actuator commands and making context inference. The proposed monitor evaluates whether the control action issued by the controller given the inferred context might result in any hazards and stops delivering unsafe commands by issuing corrective actions to mitigate hazards. We {assume} the sensor data received by the controller and the monitor are fault-free or protected using existing methods mentioned above. We also assume the monitor has a much simpler logic than the controller, so it will be easier and less expensive to be verified and made tamper-proof (e.g., using protective memories or hardware isolation \cite{Tcbase}\cite{TrustZone}). 
\label{sec:assumption}

\textbf{Context-Aware Monitoring:} A simple algorithm for the proposed monitor might involve checking the values of control commands based on ad-hoc safety rules or medical guidelines~\cite{young2018damon}. However, such a generic monitoring mechanism does not consider the current cyber-physical system status and the patient's dynamics and might incorrectly classify safe commands, leading to a large number of false alarms or missed detection and potential harm to patients~\cite{alemzadeh2016targeted}\cite{alemzadeh2013analysis}\cite{roederer2016clinician}. 

Safety, as an emergent property of CPS, is context-dependent and should be controlled by enforcing a set of constraints on the system's behavior and control actions given the current system state \cite{leveson2011engineering, yasar_ismr2019, yasar_dsn2020}. Previous works on anomaly detection in CPS have shown that considering the multi-dimensional system context, including human, cyber, and physical systems' status, leads to improved detection accuracy and latency \cite{alemzadeh2016targeted, yasar_dsn2020,lin2020challenges,sannino_mobile_2014,mcps2018}. However, most of the existing context-aware monitoring solutions rely on black-box data-driven models. Our goal is to combine expert knowledge with learning from data to improve the monitors' accuracy and transparency. 

Recent systems-theoretic approaches to safety, such as the Systems-Theoretic Accident Model and Processes (STAMP) \cite{leveson2013stpa}, propose hazard analysis methods for identifying unsafe context-dependent control actions and human-cyber-physical interactions that will lead to safety hazards. However, attempts at providing formal frameworks for models such as STAMP \cite{Asare2013}\cite{Thomas}, have still left gaps between the high-level safety requirements identified from hazard analysis and the low-level formal specification of safety properties that can be used for run-time monitoring and safety assurance. We leverage the control-theoretic notion of system context from the STAMP accident model \cite{leveson2011engineering} and develop a formal framework for the design of context-aware hazard detection and mitigation mechanisms. Our proposed framework enables the formal specification of potentially unsafe control actions given different physical contexts, which can be further refined based on simulated or real patient data to generate monitor logic. 
\section{Safety Context Specification and Learning}\label{sec:monitor-synthesis}
\vspace{-0.5em}

\begin{figure}[t!]
    \centering
    \includegraphics[width=0.82\columnwidth]{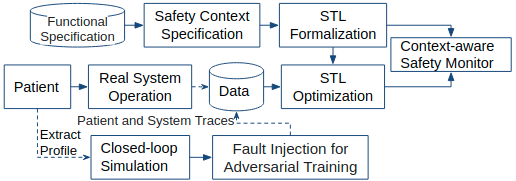}
    \vspace{-0.5em}
    \caption{Framework for Design of Context-aware Safety Monitors}
    \label{fig:designframwork}
    \vspace{-1.75em}
\end{figure}

Our overall methodology for the design of context-aware safety monitors starts with the specification of system context driven by aspects of the STAMP accident model, formalization of the context specification using STL, and its optimization through learning from system simulation traces (Fig. \ref{fig:designframwork}). 
We present a combined model and data-driven approach to provide a common framework to engineers and clinicians for the specification of safety requirements based on domain knowledge and to enable the automated refinement of safety properties to be checked at run-time using patient data. 

\vspace{-0.7em}
\subsection{Model of System Dynamics}
\vspace{-0.3em}
\subsubsection{State Space and Control Actions}
A typical MCPS controller makes an estimation of the physical system state and patient status through sensor measurements in each control loop, represented by $x_{t} = (x_{1_{t}}, \dots, x_{n_{t}}) \in \mathbb{R}^n$, where $x_{i_{t}}$ are continuous or discrete variables. At a given control cycle $t$, the controller decides on a control action, $u_t$, from a finite set of possible control actions $U = \{u_{1}, \dots, u_{r}\}$, based on current system state $x_{t}$ and sends it to the actuators. Upon execution of the issued control command by the actuators, the physical system will transition to a new state $x_{t+1}$ in the state space. 

\subsubsection{Regions of Operation}
We assume there are three mutually exclusive regions of the state space
: (i) the hazardous region ${\mathcal{X}}_{h}$, which could be further partitioned into regions associated with specific types of safety hazards $H_i$, and (ii) the safe/desirable region ${\mathcal{X}}_{*}$; and (iii) the possibly hazardous region ${\mathcal{X}}_{*<h}$, where there exists at least one control action that can move the state back to the safe region, in the absence of which the system will move to the hazardous region.

\subsubsection{Unsafe Control Action (UCA)}
A control action $u_{t}$ is unsafe if upon execution of $u_{t}$ in state $x_{t}$ the system transitions to the next state $x_{t+1}$ in the region ${\mathcal{X}}_{*<h}$ and will transit to a state in ${\mathcal{X}}_{h}$ at a future time $t' \in [t,t+T]$, where $T$ is the period that $u_{t}$ can affect the state space.


\subsection{Framework for Safety Context Specification (SCS)}
The SCS consists of two parts: (i) the UCA Specification (UCAS) that describes the system state under which a control action might be unsafe and is used by the context-aware monitor to detect UCAs and predict hazards; 
and (ii) the Hazard Mitigation Specification (HMS) that determines one or more mitigation actions to correct a potential UCA issued by the controller and prevent hazards.

\subsubsection{UCA Specification (UCAS)}
\label{subsec:UCAS}
To reduce the complexity in specifying system context for identifying UCAs, we define $\mu(x_t) = (\mu_{1}(x_t),\dots, \mu_{m}(x_t)) \in \mathbb{R}^m$, where $\mu_{i}(x_t)$ is a transformation of $x_t$, which could be the polynomial, derivative, or other possible functions of $x_t$, modeling more complex combinations of state variables and their rates of change. The set of all possible values of $\mu(x_t)$ is denoted by $\mathcal{M}$. We describe the \textit{system context} $\rho(\mu(x_t))$ as subsets of $\mathcal{M}$, defined by ranges of variables in $\mu(x_t)$, that can be mapped to the regions $\{{\mathcal{X}}_{*},{\mathcal{X}}_{*<h}, {\mathcal{X}}_{h}\}$. To identify the nature of a control action $u_t$ within a context $\rho(\mu(x_t))$, we need to determine the possibility that by issuing $u_t$ the system eventually transitions into a new context $\rho(\mu(x_{t'}))$ within the hazardous region ${\mathcal{X}}_{h}$.

An UCAS is defined as the set of all tuples in the form $(\rho(\mu(x_{t})), u_{t}, H_{i})$ such that $(\rho(\mu(x_{t})),u_{t}) \mapsto H_{i} \subset {\mathcal{X}}_{h}$ and can be generated using the following steps:
\begin{enumerate}
    \item Define the set of accidents (A) and hazards (H) of interest for the system.
    \item Identify the observable set of variables $x_{t}$ of interest related to the hazards and decide on the possible transformations $\mu(x_{t})$ and the sets $\rho(\mu(x_{t})) \in \mathcal{M}$ as completely as possible. The exact thresholds for all variables that define each subset need not be known.
    \item List all the combinations of $\rho(\mu(x_{t}))$ and $u_{t} \in U$. 
    \item Identify the combinations that might result in transitions to a hazardous region $H_{i} \subset{\mathcal{X}}_{h}$, and add tuples $(\rho(\mu(x_{t})), u_{t}, H_{i})$ into set UCAS. 
\end{enumerate}

\noindent Step 1 requires medical domain knowledge and input from clinicians. Steps 1 and 2 need to be defined manually by an analyst. Step 3 can be automated based on definitions in steps 1 and 2~\cite{Thomas}. Step 4 can be done manually, but can also be automated using dynamic modeling and simulation~\cite{asare2015thesis}.
 
\subsubsection{Hazard Mitigation Specification (HMS)} HMS is a set of tuples with the form  $(\rho(\mu(x_{t})), \boldsymbol{u}^{\rho})$, where $\boldsymbol{u}^{\rho}$ is the set of safe control actions in the context $\rho(\mu(x_{t}))$ that result in transition to ${\mathcal{X}}_{*}$. 
An HMS is generated using these steps:
\begin{enumerate}
    \item Generate the set of UCAS as above.
    \item For each context in UCAS, find all control actions $u^c_t \in U$ such that $(\rho(\mu(x_{t})),u^c_t) \mapsto {\mathcal{X}}_{*}$ and set these to $\boldsymbol{u}^{\rho}$ for that context. This step may be done manually or can be learned from simulations as well. 
\end{enumerate}


\subsection{Formalization of SCS in Temporal Logic}
For online monitoring of safety requirements and detecting UCAs, we need to describe their specification using a machine-checkable language/logic that can express complex policies in MCPS. STL is a formal specification language that is often used for rigorous specification and run-time verification of requirements in CPS \cite{bartocci_2018}. Although there has been considerable interest in using STL for specification based monitoring, most previous works relied on the specification of ad-hoc rules or clinical guidelines using STL~\cite{Eziospecification}. This paper is the first attempt to convert the high-level safety properties generated using a control-theoretic accident model into STL formalism and synthesize the generated STL formulas as an online context-aware monitor for MCPS. 

\subsubsection{Conversion of SCS to STL}
We use the bounded-time variant of STL, where all temporal operators are associated with lower and upper time-bounds. 
We refer the reader to \cite{Eziospecification} for a more detailed description of STL. Since we want our STL formula to ensure safety, we would like the formula to evaluate to true as long as a UCA is not issued in the context where it would lead to a hazard. The STL formula for a specific context, $\rho(\mu(x_{t}))$, such that $(\rho(\mu(x_{t})),u_{t}) \mapsto {\mathcal{X}}_{h}$, 
has the form:

\vspace{-1.5em}
\begin{equation}
\label{eq:eq1}
\vspace{-0.5em}
    G_{[t_{0},t_{e}]} (\varphi_{1}(\mu_{1}(x_{t})) \land \ldots \land  \varphi_{m}(\mu_{m}(x_{t})) \implies \neg u_{t})
\end{equation}

\noindent where $G$ is the globally operator $\Box$ that ensures the formula holds always during time window $[t_{0},t_{e}]$, representing the initial time and end time when we run the control system, and $(\varphi_{1}(\mu_{1}(x_{t})) \land \ldots \land  \varphi_{m}(\mu_{m}(x_{t}))$ represents the subset $\rho(\mu(x_{t}))$. Each $\varphi_{i}(\mu_{i}(x_{t}))$ is an atomic predicate that for continuous variables represents an inequality on $\mu_{i}(x_{t})$ in the form of 
$ \mu_{i}(x_{t}) \{<, \leq , >, \geq\} {\beta}_{i}$ or its combinations, where the inequality thresholds ${\beta}_{i}$ define the boundary of the subset in that dimension $\rho(\mu_{i}(x_{t}))$, and for discrete variables takes the form $(\mu_{i}(x_{t}) = \alpha_{1}) \lor \ldots \lor (\mu_{i}(x_{t}) = \alpha_{p})$, in which $\alpha_{i}$ defines a specific state or set of states.


Similarly, STL formula for HMS $(\rho(\mu(x_{t})), u^c_t)\mapsto {\mathcal{X}}_{*}$ is 
\begin{equation}
\vspace{-0.5em}
    G_{[t_{0},t_{e}]}( (F_{[0,t_{s}]}(u^c_t))\mathcal{S}(\varphi_{1}(\mu_{1}(x_{t})) \land \ldots \land  \varphi_{m}(\mu_{m}(x_{t}))))
    \label{eq:HMS}
\end{equation}

\noindent where $F$ is the eventually operator $\lozenge$ indicating $u^c_t \in \boldsymbol{u}^{\rho}$ should be taken within period $t_{s}$ since (denoted by $S$ operator) the system enters context $(\varphi_{1}(\mu_{1}(x_{t})) \land \ldots \land  \varphi_{m}(\mu_{m}(x_{t})))$. This should hold globally during $[t_{0},t_{e}]$.

The time parameter $t_s$ specifies the requirement for the latest possible time a mitigation action should be initiated after a potential UCA is detected to prevent hazards. This time is dependent on many factors, including the context $\rho(\mu(x_{t}))$ and the nature of the various safe control actions $u^c_t \in \boldsymbol{u}^{\rho}$. The specifics of determining this time requirement, in general, are beyond the scope of this paper. The estimated time between activation of a fault in the system and the occurrence of a hazard (defined as Time-to-Hazard in Section~\ref{sec:Experiment}) can provide an upper bound for specifying this time requirement.

\subsubsection{Optimization of STL Formulas}
\label{subsec:STL_learning}
The unknown boundary parameters $\beta_{i}$ in the STL formulas can be learned from actual or simulated data from the system using ML methods \cite{ml_positiveexample1}\cite{Jha2017}.
Existing STL learning approaches either rely on classification methods based on both positive and negative examples 
or use system simulation and experimentation for learning from falsification of STL properties \cite{TeLex}. In this work, we use software FI to generate hazardous data traces that potentially violate the STL formulas for SCS and use them as negative examples for learning unknown STL parameters and for adversarial training of the monitor. As shown in Fig. \ref{fig:designframwork}, patient profiles and data traces from real system operation can be used for the development of simulation models and faulty data traces and active learning in a real application.  

Given a SCS STL formula $\phi$ (Eq. \ref{eq:eq1}) and its corresponding UCAS, $(\rho(\mu(x_{t})),u_{t},H{i})$, we define an optimization problem for learning the values of the thresholds $\beta_{i}$ from a set of data traces $\mathcal{D}$. 
If the STL formula $\phi_{h}$ for UCAS (Eq. \ref{eq:eq3}), that has the same thresholds $\beta_{i}$ as $\phi$, is satisfied by a subset of hazardous traces $\mathcal{H}\subset\mathcal{D}$, the degree of satisfiability of $\phi_{h}$ for a data trace $d\in\mathcal{H}$ at time $t$ can be measured by a robustness metric $r=\mu_{i}(d(t))-\beta_{i}$ (for predicate $\mu_{i}(x_{t}) \geq\beta_{i}$). The goal of optimization is to minimize the absolute value of $r$ as a loss function over all traces in $\mathcal{H}$ to achieve tight properties \cite{jha_telex_2019}: 

\vspace{-1em}
\begin{align}
\label{eq:eq3}
    & \text{minimize} \sum_{\mathcal{H}}^{}loss(r);\ s.t.\ \\
    & r=\mu_{i}(d(t))-\beta_{i}\geqslant 0, \forall{d \in \mathcal{H}} \nonumber\\
    & \phi_{h} =\varphi_{1}(\mu_{1}(x(t))) \land \ldots \land  \varphi_{m}(\mu_{m}(d(t))) \land u_{t} \implies \lozenge {\mathcal{X}}_{h}\nonumber
\end{align}

\begin{figure}[t!]
    \centering
    \includegraphics[width=0.9\columnwidth]{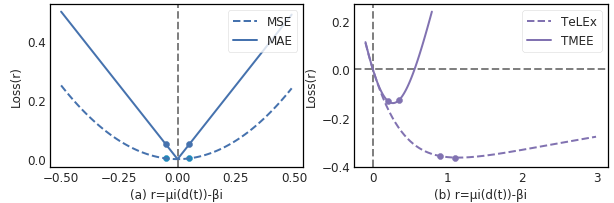}
    \vspace{-0.2em}
    \caption{Loss Functions of (a) MSE and MAE, (b) TeLEx and Our Proposed Tight Mean Exponential Error (TMEE) Function}
    \label{fig:threshold-loss}
    \vspace{-1.5em}
\end{figure}

This metric is similar to several widely-used loss functions in ML (e.g., mean squared error (MSE) and mean absolute error (MAE)) for measuring parameter estimation errors. However, as shown in Fig. \ref{fig:threshold-loss}a, when using such loss functions, the loss values can be small positive numbers near the minimum, but the actual robustness values might be small negative numbers that violate the STL formulas. A previous work, TeLEx\cite{jha_telex_2019}, addressed this problem by introducing a tightness function \cite{TeLex} to measure loss (Fig. \ref{fig:threshold-loss}b), but the thresholds learned using such a loss function are not tight enough without manually adjusting. In this paper, we introduce a Tight Mean Exponential Error (TMEE) loss function, as shown below: 
\begin{equation}
\label{eq:loss_th}
    loss(r)=E[e^{-r}+r-\frac{1}{1+e^{-2r}}],\ r=\mu_{i}(d(t))-\beta_{i}
\end{equation}
which learns tight thresholds while ensuring that safety specification STL formulas are not violated by normal data traces. 

We used an extension of the Limited-memory Broyden-Fletcher-Goldfarb-Shanno algorithm called L-BFGS-B \cite{BFGS_Morales}, an optimization algorithm in the family of quasi-Newton methods for parameter estimation. Unlike typical quasi-Newton methods \cite{Quasi-Newton} that calculate the inverse of the Hessian matrix directly, we used two-loop recursion \cite{two_loop_recu} to estimate it. The L-BFGS-B algorithm then used the gradient of the loss function and the estimated inverse Hessian matrix to guide the optimization. 

Our preliminary experiments of learning thresholds for a population-based monitor using the TeLEx loss function could not always converge to a solution. Also, the final context-aware monitor achieved lower accuracy than a monitor with tight thresholds learned using our optimization approach. 

For the synthesis of a context-aware mitigation mechanism based on HMS (Equation (\ref{eq:HMS})), we need to refine our STL learning method to learn the unknown time parameter $t_{s}$ in addition to safety thresholds $\beta_{i}$.
In this paper, we mainly focus on the evaluation of hazard prediction and use a fixed non-context-dependent mitigation algorithm for a fair comparison among different monitors (see Algorithm \ref{alg:mitigationalgorithm} in Section \ref{sec:mitigationdesc}).

\vspace{-0.5em}
\section{Case Study of Artificial Pancreas Systems}
\begin{figure}[b]
    \vspace{-1.5em}

    \centering
    \includegraphics[width=0.9\columnwidth]{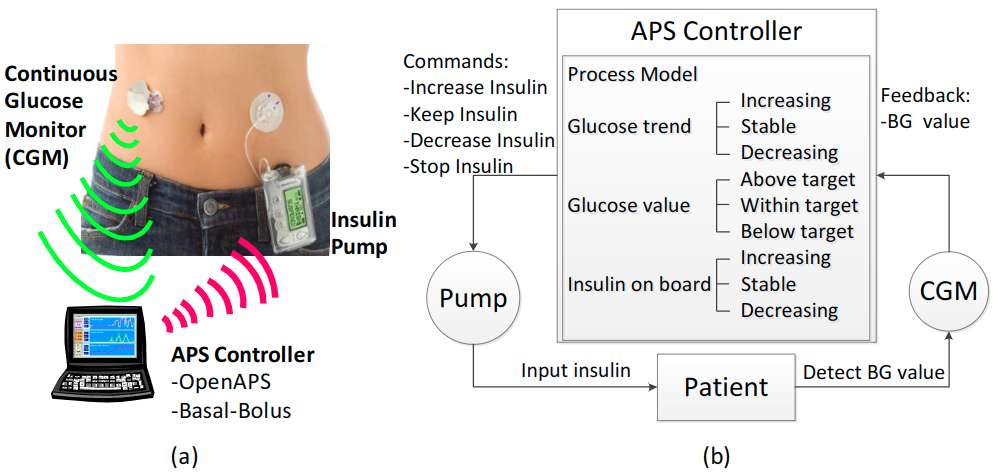}
    \vspace{-0.5em}
    \caption{(a) Artificial Pancreas System, (b) A Typical APS Controller.}
    \label{fig:controller-closeloop}
\end{figure}

\vspace{-0.5em}
To evaluate the effectiveness of our approach, we applied our methodology to the case of developing run-time monitors for Artificial Pancreas Systems (APS). APS are responsible for regulating Blood Glucose (BG) dynamics by monitoring BG concentration in the patient's body through sensor data collected from a Continuous Glucose Monitor (CGM) and providing the right insulin rate to the patient through a pump (Fig. \ref{fig:controller-closeloop}a). The control software estimates the current patient status (e.g., BG value, Insulin on Board (IOB)) and calculates the next recommended insulin value for the patient (Fig.~\ref{fig:controller-closeloop}b). 

The U.S. Food and Drug Administration (FDA) recommends that APS should be able to adequately mitigate the risks associated with erroneous readings by the CGM sensors and inappropriate doses delivered by the insulin pump~\cite{us2012content}. It also suggests the simulation and evaluation of the impact of such errors during the device development process.

\vspace{-0.5em}
\subsection{Closed-loop Simulation}
\label{subsec:closed_loop}
\vspace{-0.5em}

To evaluate the effect of the system on patients through simulation, we developed a closed-loop simulation testbed by integrating two widely-used APS controllers with patient glucose simulators. Our main case study is the OpenAPS \cite{openSourceOpenAPS} control software with the Glucosym patient simulator \cite{openSourceGlucosym} (Fig. \ref{fig:FI-framework}a). The Glucosym simulator contains patient models derived from data collected from 10 actual adult patients with Type I diabetes mellitus 
aged 42.5 $\pm$ 11.5 years \cite{KanderianT1}. 
To further test the generalization of the proposed approach, we also used the state-of-the-art UVA-Padova Type 1 Diabetes Simulator S2013 (T1DS2013) \cite{man2014uva}, which contains 30 virtual patients and has been accepted by the FDA for pre-clinical testing, together with a Basal-Bolus controller \cite{BBcontroller}.

 \vspace{-0.5em}
\subsection{Safety Context Specification (SCS)}
\label{sec:casestudy}
\vspace{-0.5em}


We first identified the set of accidents and the safety hazards that might happen due to UCAs issued by an APS controller. 

\textbf{Accidents:} In Type I diabetes, which the APS is designed for, there are two main accidents that we are concerned about:
\begin{itemize}
\item \textbf{A1:} Complications from hypoglycemia (BG level too low), including seizure, loss of consciousness, and death.
\item \textbf{A2:} Complications from hyperglycemia (BG level too high), including tissue damage and morbidities such as retinopathy and in extreme cases, death \cite{Complicationshyperglycemia}. 
\end{itemize}

\textbf{Hazards:} The set of system states under the control of the APS that together with the other conditions might lead to accidents include: 
\begin{itemize}
    \item \textbf{H1:} Too much insulin is infused, which will reduce the BG and might lead to \textbf{A1}.
    \item \textbf{H2:} Too little insulin is infused, which causes the BG to increase and could lead to \textbf{A2}.
\end{itemize}

 
We then identified some transformations of interest on $x_{t} = (BG_t)$ as $\mu(x_t) = (BG_t, dBG_t/dt, IoB_t, dIoB_t/dt)$, including both the state variable $BG_t$ and its rate of change as well as estimated IoB and its rate of change, which can be calculated based on previous insulin deliveries. Then following steps in Section \ref{subsec:UCAS}, the formalized UCAS for APS was generated by identifying the combinations of specific ranges in $\mu(x_t)$ and insulin control commands $u_t \in \{u1, u2, u3, u4\}$ (as shown in Table \ref{table:stltable}) that can potentially be hazardous. Each row in Table \ref{table:stltable} shows STL formulas to be checked by the monitor. For example, the first row is the formal representation of a UCAS, $(\rho(\mu(x_{t}))=$ (BG>BGT, BG'>0, IOB'<0, IOB<$\beta_{1}$), $u_{t}=u_{1})\mapsto H2 \subset {\mathcal{X}}_{h}$, requiring that under such a system context $\rho(\mu(x_{t}))$, the UCA $u_{1}$ (decrease insulin) should not be issued at anytime during $[t_{0},t_{e}]$. Otherwise, an H2 hazard might happen.
\begin{table}[b!]
\centering
\vspace{-1.5em}
\caption{Safety Context Specification for APS Described in STL}
\vspace{-0.5em}
\label{table:stltable}
\resizebox{\columnwidth}{!}{%
\begin{threeparttable}

\begin{tabular}{|c|l|c|} \hline
\textbf{Rule} & \multicolumn{1}{c|}{\textbf{STL Description}} &\textbf{Hazard Type} \\ 
\textbf{No.} && (if violated) \\ \hline

1 & $G_{[t_{0},t_{e}]}$((BG>BGT$\wedge$ BG'>0)$\wedge$ (IOB'<0$\wedge$ IOB<$\beta_{1}$)$\Rightarrow\neg$ $u_{1}$) & H2 
\\ \hline
2 & $G_{[t_{0},t_{e}]}$((BG>BGT$\wedge$ BG'>0)$\wedge$ (IOB'=0$\wedge$ IOB<$\beta_{2}$)$\Rightarrow\neg$ $u_{1}$) & H2 
\\ \hline
3 & $G_{[t_{0},t_{e}]}$((BG>BGT$\wedge$ BG'<0)$\wedge$ (IOB'>0$\wedge$ IOB<$\beta_{3}$)$\Rightarrow\neg$ $u_{1}$) & H2 
\\ \hline
4 & $G_{[t_{0},t_{e}]}$((BG>BGT$\wedge$ BG'<0)$\wedge$ (IOB'<0$\wedge$ IOB<$\beta_{4}$)$\Rightarrow\neg$ $u_{1}$) & H2 
\\ \hline
5 & $G_{[t_{0},t_{e}]}$((BG>BGT$\wedge$ BG'<0)$\wedge$ (IOB'=0$\wedge$ IOB<$\beta_{5}$)$\Rightarrow\neg$ $u_{1}$) & H2 
\\ \hline
6 & $G_{[t_{0},t_{e}]}$((BG<BGT$\wedge$ BG'<0)$\wedge$ (IOB'>0$\wedge$ IOB>$\beta_{6}$)$\Rightarrow\neg$ $u_{2}$) & H1 
\\ \hline
7 & $G_{[t_{0},t_{e}]}$((BG<BGT$\wedge$ BG'<0)$\wedge$ (IOB'<0$\wedge$ IOB>$\beta_{7}$)$\Rightarrow\neg$ $u_{2}$) & H1 
\\ \hline
8 & $G_{[t_{0},t_{e}]}$((BG<BGT$\wedge$ BG'<0)$\wedge$ (IOB'=0$\wedge$ IOB>$\beta_{8}$)$\Rightarrow\neg$ $u_{2}$) & H1 
\\ \hline
9 & $G_{[t_{0},t_{e}]}$((BG>BGT$\wedge$ IOB<$\beta_{9}$)$\Rightarrow\neg$ $u_{3}$) & H2
\\ \hline
10 & $G_{[t_{0},t_{e}]}$((BG<$\beta_{21}$)$\Rightarrow$ $u_{3}$) & H1 
\\ \hline
11 & $G_{[t_{0},t_{e}]}$((BG>BGT$\wedge$ BG'>0)$\wedge$ (IOB'<=0$\wedge$ IOB<$\beta_{10}$)) $\Rightarrow\neg$ $u_{4}$) & H2 
\\ \hline
12 & $G_{[t_{0},t_{e}]}$((BG<BGT$\wedge$ BG'<0)$\wedge$ (IOB'>=0$\wedge$ IOB>$\beta_{11}$)$\Rightarrow\neg$ $u_{4}$) & H1\\

\hline

\end{tabular}

\begin{tablenotes}\footnotesize
\item[*] BGT: BG target value; 
$BG' =dBG/dt$, $IOB'=dIOB/dt$.
\item[*] $t_{0},t_{e}$: start time and end time of the simulation; 
\item[*] $u_{1,2,3,4}:$decrease\_insulin, increase\_insulin, stop\_insulin, keep\_insulin.
\end{tablenotes}

\end{threeparttable}
}
\end{table}
Here the boundary threshold $\beta_{1}$ for the estimated IOB is learned from data. 

It should be noted that the generated UCAS and the final monitor logic can be used for different APS controllers that share the same functional specification. 

\vspace{-0.7em}
\subsection{Safety Context Learning}
\label{sec:labeling}
\vspace{-0.3em}
\subsubsection{Adversarial Training using Fault Injection}
We use software FI for generating hazardous data traces that can potentially violate the SCS formulas shown in Table \ref{table:stltable} and use them as negative examples for learning the boundary thresholds $\beta_i$. Specifically, as shown in Fig. \ref{fig:FI-framework}a, we inject a diverse set of faults inside a closed-loop APS controller with glucose simulator and use the generated faulty data for adversarial training and testing of the context-aware monitors as well as other baseline monitors.

\textbf{Threat Model:} We assume that both accidental faults or attacks, similar to those reported for CPS/APS (see Table \ref{tab:threat_model}), can target the APS controller and, once activated/initiated, can manifest as errors in inputs, outputs, and the internal state variables of the APS control software and cause the hazard types defined in Section \ref{sec:casestudy}. 
So our source-level FI engine directly perturbs the values of the controller's state variables within the acceptable range over a period to simulate the effect of such errors. We assume errors are transient and only occur once for a particular duration per simulation.

For malicious attacks, we assume that attackers have obtained unauthorized remote access \cite{FDApumprecall} to an APS control system by exploiting weaknesses such as stolen credentials\cite{Stuxnet}, vulnerable services \cite{Zetterhackhospital}, or insider attacks to penetrate the network \cite{alemzadeh2016targeted} that the target APS controller connects to. Even for an APS control system that does not connect to a network, the attacker can still use a USB port or Bluetooth to get access and deploy malware. 


\label{subsec:threat_model}
\begin{table}[t]
    \centering
    \caption{Simulated Fault and Attack Scenarios}
    \vspace{-0.5em}
    \resizebox{\columnwidth}{!}{
    \begin{tabular}{|l|l|l|}
    \hline
    \textbf{Type}   & \textbf{Approach}& \textbf{Simulated Scenario} \\ \hline
    
    Truncate & Change output variables to zero value   \cite{li2011hijacking}\cite{ramkissoon2017review}& Availability attack  \cite{availabilityattack}        \\ \hline
    
    Hold            & Stop refreshing selected input/output variables & DoS attack \cite{dosattack}\cite{Bonacidos}\\ 
    
    &\cite{attack_pcs}\cite{ramkissoon2017review} & \\\hline
    
    Max/min & Change the value of targeted variables to their& {Integrity attack \cite{li2011hijacking}\cite{FDApumprecall}/ } \\ 
    
    &maximum or minimum allowed values  \cite{attack_pcs} \cite{jha2019mlbased}&\\ \cline{1-2}
    
    Add/Sub         & Add or subtract an arbitrary or particular value& {Memory fault}        \\ 
    
    & to a targeted variable  \cite{attack_pcs}\cite{sadhu2020onboard}& \\\hline
    \end{tabular}
    }
    \label{tab:threat_model}
    \vspace{-2em}
\end{table}



\subsubsection{Labeling Hazards using BG Risk Index}
\label{sec:risk_index}
To label data points as normal or hazardous in our simulation traces, we exploited the notion of the Risk Index (RI) \cite{clarke2009statistical,kovatchev2017metrics} that captures both the glucose variability and its associated risks for hypo- and hyperglycemia. First we calculated the BG risk function for each BG reading as follows: 
\begin{equation}
   risk(BG) = 10*(1.509*[(ln(BG))^{1.084} - 5.381])^2
\end{equation}

\noindent Then the left and right branches of the BG risk function ($risk(BG) < 0$ and $risk(BG) > 0$) were separated to calculate low (LBGI) and high (HBGI) BG risk indices for a window of BG readings by taking average risk index on each side.
We then marked a window (e.g., one hour) of BG readings as hazardous if the risk indices LBGI or HBGI for that window crossed a high-risk threshold\footnote{$LBGI > 5$ and $HBGI > 9$ as defined by previous works~\cite{kovatchev2017metrics}\cite{Kovatchev1870} \vspace{0em}} and kept increasing, indicating a high chance of hypo- or hyperglycemia. An example of a labeled simulation trace is shown in Fig. \ref{fig:FI-framework}b.

\vspace{-0.5em}
\subsection{Hazard Mitigation and Recovery}
\label{sec:mitigationdesc}
When the monitor detects a UCA is issued by the controller, it will try to mitigate the potential hazards by correcting the command (regardless of its value being "out-of-the-range" or not) and delivering a new command ($u^c_t\in \boldsymbol{u}^\rho$) to the actuator. For example, it can decrease the insulin when it is more than needed, or add suitable insulin when the provided command is insufficient. The correction of a UCA will continue until the system moves back to a safe state and the monitor stops raising alerts. Designing a mitigation mechanism with a high hazard recovery rate while introducing as few new hazards as possible is a challenging task. Algorithm \ref{alg:mitigationalgorithm} shows one possible implementation of a mitigation algorithm to prevent hazards in APS. Further investigation of mitigation algorithms based on formal specification and learning from simulation data is beyond this paper's scope. 
\begin{algorithm}[t!]
 \footnotesize
 \DontPrintSemicolon
    \caption{Hazard Mitigation Algorithm}
    \label{alg:mitigationalgorithm}
    Mitigate $\gets$ 0\\
    \While{$t<t_{e}$}
    {
         $\mu(x_{t}) \gets (BG_{t},IOB_{t},BG^{'}_{t},IOB^{'}_{t})$\; 
         $u_{t},u^c_t \gets u_{i} \in \{u1, u2, u3, u4\}$\;

         \lIf{$\rho(\mu(x_{t})) \in {\mathcal{X}}_{*}$}{Mitigate $\gets$ 0, \textbf{continue}}
        
         \For{$\phi_i$ in STL of SCS}{
          \uIf {($\rho(\mu(x_{t})), u_{t}) \textit{ violates } \phi_i$}
        {
            Mitigate$\gets$1, \textit{Hazard} $\gets H_{i} \in \{H1, H2\} $\\
        }         
        }

        \uIf{Mitigate == 1}
        {

            \lIf{$\textit{Hazard} == H_{1}$}
            {
                $u^c_t \gets 0$ 
            }
            \lElseIf{$\textit{Hazard} == H_{2}$}
            {
                $u^c_t \gets f(\rho(\mu(x_{t})), u_{t}) \in \boldsymbol{u}^{\rho}$
            }
        }
    }
    \algorithmfootnote{
    $f(.)$ describes a context-dependent function for selecting the mitigation action. In our experiments, we instead use a fixed maximum value of insulin to enable a fair comparison with baseline non-context-aware monitors. \vspace{-2.5em}}
\end{algorithm}

\vspace{-0.5em}
\section{Experimental Evaluation}
\vspace{-0.3em}
\label{sec:Experiment}
\begin{figure*}[t!]
    \centering
    \includegraphics[width=0.81\textwidth]{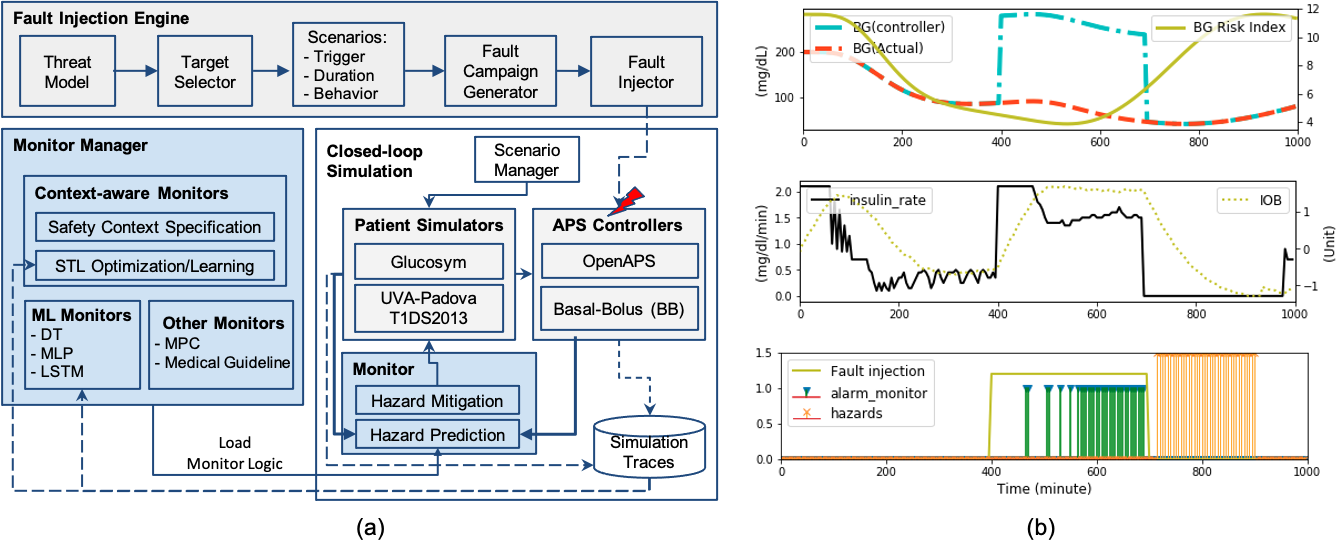}
    \vspace{-1em}
    \caption{(a) Experimental Setup for Evaluation of Different Safety Monitors, integrating Two Closed-loop Simulation Platforms (Glucosym Simulator with OpenAps Controller and UVA-Padova T1DS2013 Simulator with Basal-Bolus Controller), and Software FI Engine. (b) An example Simulation Trace with Injected Faults and Labeled Hazards}
    \vspace{-1.75em}
    \label{fig:FI-framework}
\end{figure*}

Fig. \ref{fig:FI-framework}a shows our overall experimental setup that integrates the closed-loop simulation of the APS control systems with a software FI engine to evaluate different safety monitors. This open-source simulation environment is publicly available\footnote{ https://github.com/UVA-DSA/ContextSafetyMonitorAPS
\vspace{0em}}.We ran the OpenAPS controller with the Glucosym simulator on a virtual machine running Ubuntu 14.04 LTS. The experiments with the T1DS2013 simulator were conducted on an x86\_64 PC with an Intel Core i5 CPU @ 3.20GHz and 16GB RAM, running Linux Ubuntu 16.04 LTS.  We used TensorFlow \cite{tensorflow} v.2.0.0 to train our ML models and Scikit-learn \cite{Scikit-learn} v.0.22.2 for data pre-processing and experimental evaluation.

\vspace{-0.5em}
\subsection{Patient Simulations}
\vspace{-0.5em}
In each experiment, we had the patient simulator interacting with the APS controller (OpenAPS or Basal-Bolus controller \cite{VANBRUNT2013809}) for 150 iterations (about 12 hours), with each step/iteration in the simulation representing 5 minutes in the real APS control system. 
Simulations began with the patient at an initial glucose value between 80 and 200 mg/dl. We assumed the patient had no meals or exercise during the simulation period, mimicking a scenario of patient eating dinner, going to sleep, and having the next meal after our simulation period the next day. 
To account for some inter-patient variability, each system version (without a safety monitor and with each different safety monitor) was evaluated using 20 different patient profiles (10 patients in the Glucosym simulator and 10 in the T1DS2013 simulator). We also explored seven different initial glucose values for each patient.
\vspace{-0.5em}
\subsection{Fault Injection Experiments}
\vspace{-0.3em}
For each FI scenario shown in Table \ref{tab:threat_model}, the FI engine selected (a) the name of the target state variable, (b) the injected error value, (c) the activation condition (trigger), and (d) the injection duration. For each scenario, we randomly chose from 9 different start times and durations to inject the fault. The combination of these parameters resulted in a total of 882 fault injections for every patient scenario. That translated into a total number of 2,646,000 simulation samples used for training and testing different monitors.

We used the data from all the patients with a 4-fold cross-validation setup for threshold learning and evaluation of our context-aware with refined threshold (CAWT) monitor as well as model training and testing of the ML baseline monitors.

\subsection{Baseline Monitors}
We compared the proposed CAWT monitor's performance in accurate and timely prediction of hazards to the following baseline monitors, representative of existing safety monitoring solutions for MCPS and APS as proposed by previous works.

\subsubsection{Medical Guidelines Monitor}

We used the data authenticity monitor proposed in \cite{young2018damon} as a baseline monitor designed based on the generic medical guidelines without any knowledge of the APS controller or the patient characteristics (referred to as Guideline). The safety rules of the Guideline monitor are shown in Table \ref{table:baselineRules}. They state that the BG value should maintain in a normal range [70, 180] mg/dL and should not change too fast. If BG is lower than its tenth percentile $\lambda_{10}$ or higher than its ninetieth percentile $\lambda_{90}$, an APS controller should bring it back to a safe range within $\alpha$ (e.g., 25) minutes. 


\begin{table}[b]
\vspace{-1.5em}
\begin{center}
\caption{Rules of Medical Guideline Monitor}
\vspace{-0.75em}
\label{table:baselineRules}
\resizebox{0.35\textwidth}{!}{%
\begin{tabular}{ | c | l |} \hline
    \textbf{No.} & \textbf{Description} \\ \hline
    1 & $\phi1 = \Box(BG>70)\wedge(BG<180))$\\ \hline
    2 & $\phi2 = \Box((\Delta BG> -5$$)\wedge($$\Delta BG < 3))$\\ \hline
    3 & $\phi3 =((BG< \lambda_{10})\Rightarrow \lozenge_{[0,\alpha]}(BG > \lambda_{10}) )$ \\ \hline
    4 & $\phi4 =((BG> \lambda_{90})\Rightarrow \lozenge_{[0,\alpha]}(BG < \lambda_{90}) )$ \\ \hline
\end{tabular}}
\end{center}
\end{table}

\subsubsection{Model Predictive Control Monitor}
Another baseline monitor that we considered was the widely-used Model Predictive Control (MPC) monitor~\cite{MPCCairoli2019,MPCRaman2014}.  We used the Bergman \& Sherwin model\cite{KanderianT1} for this monitor:
\begin{equation}
\vspace{-0.5em}
\Scale[0.92]{dBG(t)/dt=-(GEZI+I_{EFF})*BG(t)+EPG+R_{A}(t)}
\end{equation}
where, 
$GEZI$ characterizes the effect of glucose per se to increase glucose uptake into cells and lower endogenous glucose production at zero insulin; $EGP$ is the endogenous glucose production rate that would be estimated at zero insulin; $I_{EFF}$ is insulin effect; and $R_{A}(t)$ represents glucose appearance following a meal. The MPC monitor estimates the possible BG value ($BG_{t+1}$) after executing the pump's command ($I_t$) on the patient's current state ($BG_t$). If the predicted BG value goes beyond the patient's normal range ([70,180] mg/dL as defined by the medical guidelines), an alarm will be generated.

\subsubsection{Context-aware Monitor without Threshold Learning} 
\label{subsec:CAWT-Generic} We also designed a context-aware baseline monitor with the same STL logic as the proposed CAWT monitor (see Table \ref{table:stltable}) but without learning the thresholds through data-driven STL optimization described in Section \ref{subsec:STL_learning}. We refer to this baseline monitor as the CAWOT monitor.

\subsubsection{ML-based Monitors}

We used the widely-used ML approaches, Decision Trees (DT), Multi-layer Perceptron (MLP), and Long-Short Term Memory (LSTM), to train three baseline monitors, representative of ML-based monitors previously proposed in~\cite{yasar_dsn2020}.
For DT and MLP, we model the task of detecting UCA as a context-specific conditional event, as shown below: 
\vspace{-0.8em}
\begin{equation}
\label{eq:ml-model}
\begin{split}
    x_{t} &= (x_{1_{t}},x_{2_{t}},...,x_{n_{t}}) \\
    y_{t} &= p(\exists t'\in [t,t_{e}]: x_{t'} \in {\mathcal{X}}_{h}|x_{t},u_{t})
\end{split}
\end{equation}
\vspace{-1.2em}

The input is the current system state $x_t$ and the issued control action $u_{t}$ and the output $y_{t}$ is a binary classification of $u_t$ to safe or unsafe. For training the model, the output was labeled as positive for a given input if any hazard happened at a future time $t'\in[t,t_{e}]$. We marked a simulation as hazardous if any sample within the simulation was unsafe. We used two fully connected layer MLP, comprising 256 and 128 neutrons, followed by a fully connected layer with ReLU activation and a final softmax layer to obtain the hazard probabilities. 

We also built an LSTM model as a baseline monitor because of its advantage in handling time-series data.
For the LSTM, we used input data with a sliding time-window of \textit{k}:
\begin{equation}
\label{eq:lstm-model}
\begin{split}
    X_{t} &= (x_{t},x_{t+1},...,x_{t+k}),  U_{t} =(u_{t},u_{t+1},...,u_{t+k})\\
    y_{t} &= p(\exists t'\in [t,t_{e}]: x_{t'} \in {\mathcal{X}}_{h} |X_{t},U_{t})
\end{split}
\end{equation}
We experimented with different model architectures, and the best model we got was a two-layer stacked LSTM 
with an input time step of 6 (meaning 30 minutes data), consisting of 128 and 64 LSTM units, respectively, followed by a fully connected layer with softmax activation. 
We trained the MLP and LSTM models using the Adam \cite{kingma2017adam} optimizer with the loss function of sparse categorical cross-entropy and a learning rate of 0.001. To avoid over-fitting, we added dropout regularization and early stopping on a held-out validation set.

\vspace{-1.0em}
\subsection{Metrics}
\vspace{-0.5em}
\label{subsec:metrics}
We introduce the following metrics for the evaluation of system resilience and performance of safety monitors:


\begin{itemize}[leftmargin=*]
    \item \textbf{\textit{Hazard Coverage}} is defined as the conditional probability that given activation of a safety-critical fault in the system by FI, it leads to an unsafe system state or a hazard.
    \item \textbf{\textit{Time-to-Hazard (TTH)}} measures the time between activation of a fault ($t_f$) to occurrence of a hazard ($t_h$) (Fig.~\ref{fig:metric}). 
    \item \textbf{\textit{Prediction Accuracy}} represents the performance of the safety monitors in accurate prediction of hazards, measured using false positive rate (FPR), false negative rate (FNR), accuracy (ACC), and F1 score.
     \begin{itemize}
         \item \textbf{\textit{Sample Level with Tolerance Window:}} Using the traditional point-wise binary classification metrics, an FP is declared for all the samples in a simulation where the monitor detects a hazard and the ground truth indicates no hazard. But for hazard \textit{prediction}, it is desirable that a monitor generates alerts \textit{before} a hazard happens. So we adopt a modified version of standard classification metrics \cite{Classification-Tharwat}, proposed for sequential data \cite{scharwachter2020statistical}\cite{tatbul2018precision}\cite{range2020zhou}, where 
         a tolerance window before the start time of hazard ($t_h$) is used for calculation of the metrics (see Fig. \ref{fig:metric}). Table \ref{tab:confusionmatrix} shows the confusion matrix with a tolerance window.
         \item \textbf{\textit{Simulation Level with Two Regions:}} Considering the whole data trace of a simulation as a single case, we also calculate accuracy at the simulation level. In that case, however, a TP is declared whenever an alert is generated during a hazardous data trace regardless of when hazards happen. Thus, for simulation level evaluation, we divide a data trace into two regions based on the time of activation of a fault ($t_f$) ($[0,t_{f}]$ and $[t_{f},t_{e}]$ in Fig. \ref{fig:metric}), and calculate the classification metrics separately for each region.
     \end{itemize}
    
   
    
\begin{figure}[b]
\vspace{-1.75em}
    \centering
    \includegraphics[width=.7\columnwidth]{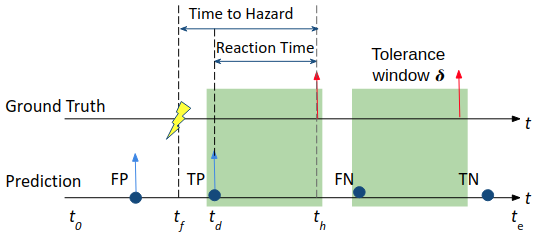}
\vspace{-0.5em}    
    \caption{Hazard Prediction Accuracy with Tolerance Window $\delta$ (green area): TP: Hazard (red arrow) occurs no latter than $\delta$ after a prediction (blue arrow); FP: No hazard happens in [0,$\delta$] after an alert; FN: Hazard occurs without a prediction in the window $\delta$ ahead; TN: No hazard happens in [0,$\delta$] after a negative prediction.}
    \label{fig:metric}
\end{figure}

     
     \item \textbf{\textit{Reaction Time}} is the time difference between a monitor alert ($t_d$) and the occurrence of a hazard ($t_h$) (Fig.~\ref{fig:metric}). This is the maximum time we have for taking any mitigation action before the hazard happens, with positive values representing \textit{early detection}, and measures the timeliness of the monitor.
     
     \item \textbf{\textit{Recovery Rate}} is the percentage of potential hazards that are prevented by the safety monitor's mitigation strategy and is affected by both the prediction accuracy and latency.

    \item \textbf{\textit{Average Risk}} is a metric for assessing the impact of monitor performance on patient safety by considering the \textit{consequences} of both FP and FN cases and the possibility of harm to patient. FNs put the patient in a hazardous situation without any warning or mitigation, and FPs not only bother the patient with unnecessary alerts but might also cause new hazards after needless mitigation. It is defined as follows:

\vspace{-0.5em}
\begin{equation}
\label{eq:avgrisk1}
    Risk_{avg}= \frac{1}{N}[\sum\ _{i=1}^{N_{FN}}\bar{RI}(i)+\sum\ _{i=1}^{N'_{P}}\bar{RI}(i)]
\vspace{-0.25em}
\end{equation}
where, $\bar{RI}(i)$ is the average risk index (for APS, defined as BG Risk Index in Section \ref{sec:labeling}) of $ith$ simulation, $N$ is the total number of simulations, $N_{FN}$ is the number of FN cases, and $N'_{P}$ is the number of new hazards that are introduced by mitigation of FP cases.

\end{itemize}


\begin{table}[t]
\centering

\caption{Confusion Matrix for Sequential Data with Tolerance Window $\delta$, Modified from \cite{scharwachter2020statistical}}
\label{tab:confusionmatrix}

\vspace{-0.5em}

\resizebox{\columnwidth}{!}{%
\begin{threeparttable}

\begin{tabular}{|l|l|l|}
\hline
 & Ground Truth Positive & Ground Truth Negative \\ \hline
PP &  $\sum_{t'=t-\delta'_{t}}^{t}P(t')>0 \&\& \sum_{t'=t}^{t+\delta}G(t')>0$ &  P(t)>0 $\&\& \sum_{t'=t}^{t+\delta}G(t')==0$\\\hline

PN &$\sum_{t'=t-\delta'_{t}}^{t}P(t')==0 \&\& \sum_{t'=t}^{t+\delta}G(t')>0$  & P(t)==0 $\&\& \sum_{t'=t}^{t+\delta}G(t')==0$\\\hline

\end{tabular}%

\begin{tablenotes}\footnotesize
\item[*] PP: Predicted positive; PN: Predicted negative; P(t)/G(t): Prediction/Ground truth at time t; 
$t-{\delta}'_{t}$: Start time of a window $\delta$, ending with a positive ground truth, that includes t.
\end{tablenotes}

\end{threeparttable}

}
\vspace{-2.5em}
\end{table}

\vspace{-0.7em}
\subsection{Results}
\label{subsec:results}
\vspace{-0.3em}
\subsubsection{Resilience of Baseline APS without Safety Monitor}
We first analyzed the resilience of the baseline OpenAPS software, which is already designed with safety features such as a maximum threshold and an auto-adjusted control algorithm \cite{apscontrol}, without any safety monitors in the presence of faults.

\begin{figure}[b]
    \vspace{-1.5em}
    \centering
    \includegraphics[width=0.8\columnwidth]{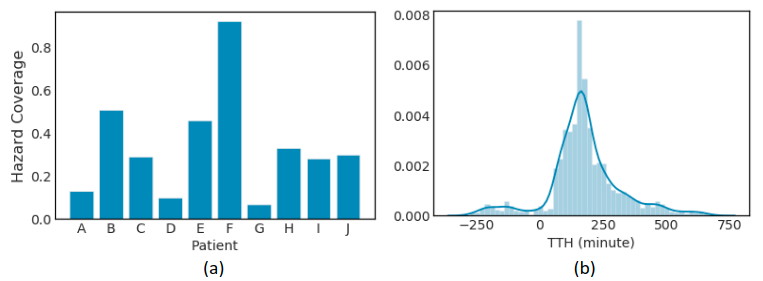}
    \vspace{-0.5em}
    \caption{(a) Hazard Coverage; (b) Time to Hazard (TTH) Distribution}
    \label{fig:hazardcoverage-mtth}
\end{figure}

\textbf{Effectiveness of FI:} 
Experimental results showed that our FI could achieve an overall 33.9\% hazard coverage on the Glucosym simulator, which reflects our FI engine's efficiency in introducing enough faulty data for adversarial training as well as OpenAPS's inadequacy in tolerating safety-critical faults and attacks. 
However, as shown in Fig. \ref{fig:hazardcoverage-mtth}a, the hazard coverage was quite different across different patient profiles, ranging from 6.7\% to 92.4\% across ten patients. This shows some evidence on the importance of specifying patient-specific safety requirements for the design of monitors. 

\textbf{OpenAPS Resilience:}
We further evaluated the resilience of OpenAPS using the TTH metric. We analyzed the distribution of this metric (Fig.~\ref{fig:hazardcoverage-mtth}b) to help with the specification of time requirements for hazard prediction and mitigation. 
Fig. \ref{fig:hazardcoverage-mtth}b shows an average TTH of about 3 hours based on all the simulation data. It should be noted that the human body has a considerable lag and is a slow dynamic system, and it usually takes hours for the BG to transmit into the vessel and for insulin to take effect. 
Besides, the TTH in 7.1\% of hazardous simulations was less than zero, which means that the hazards happened even before we injected any faults to the controller, indicating the inadequacy of the APS control algorithm.





\begin{figure}[t!]
    \centering
    \includegraphics[width=0.9\columnwidth]{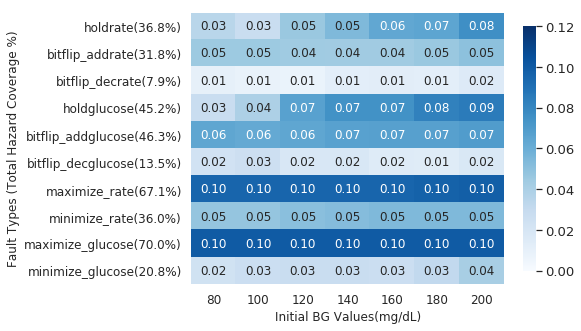}
    \vspace{-0.5em}
    \caption{Average Hazard Coverage with Different Fault Types and Initial BG Values on Glucosym}
    \label{fig:hazardcoverage_heatmap}
    \vspace{-2em}
\end{figure}

\textbf{Fault Types:} We also analyzed the relationship between the overall hazard coverage (averaged across all the patients) with the fault types and initial BG values. As shown in Fig. \ref{fig:hazardcoverage_heatmap}, we observed that for the set of patient profiles that we studied, OpenAPS controller was more vulnerable to \textit{maximize\_rate} and \textit{maximize\_glucose} attacks or faults while less vulnerable to \textit{bitflip\_decrate} or \textit{bitflip\_decglucose} faults. This is because in the latter cases, the APS controller can inject extra insulin after the faults go away to avoid high-risk situations. In contrast, a large amount of IOB remains in the body, even after the effects of the former faults or attacks have disappeared, and keeps decreasing BG and puts patients at the risk of hypoglycemia. Further, we observed an increase in hazard coverage when the initial BG values increased in about half of the fault types, which indicates faults may have more impact on risky patients.


\subsubsection{Monitor Prediction Accuracy}

Table \ref{tab:summary} shows the average performance of the CAWT monitor (over all the patients and fault scenarios) versus all other non-ML-based baseline monitors. 
We see that for the same number of simulations, the proposed CAWT monitor outperformed the Guideline monitor and MPC monitor in every metric listed in Table \ref{tab:summary} on the Glucosym simulator. Even though the CAWT monitor had a slightly larger FNR on the T1DS2013 simulator, it held the lowest FPR and achieved the highest overall accuracy and F1 score. We will further analyze the trade-off between low FPR and low FNR as well as their average risk in Section \ref{sec:mitigation-result}.


Without learning the thresholds of BG values and IOB, the CAWOT monitor had a higher FPR and lower accuracy and F1 score than the MPC monitor on the T1DS2013 simulator, but still kept the advantage over Guideline and MPC monitors on the Glucosym simulator, which demonstrates the benefit of knowing the context as well as the disadvantage of not specifying boundary thresholds in SCS. Considering its worse performance than the CAWT monitor, we do not show the CAWOT monitor's results in the following sections. 




\begin{table}[t!]
\centering
\caption{Performance of CAWT Monitor vs. Non-ML Monitors}
\vspace{-0.75em}
\label{tab:summary}
\resizebox{\columnwidth}{!}
{


\begin{tabular}{|c|l|l|l|l|l|l|l|}
\hline
\multicolumn{1}{|l|}{\multirow{2}{*}{\textbf{Simulator}}} & \multirow{2}{*}{\textbf{Monitor}} & \multirow{2}{*}{\textbf{\begin{tabular}[c]{@{}l@{}}No. Sim. \end{tabular}}} & \multirow{2}{*}{\textbf{Hazard\%}} & \multirow{2}{*}{\textbf{FPR}} & \multirow{2}{*}{\textbf{FNR}} & \multirow{2}{*}{\textbf{ACC}} & \multirow{2}{*}{\textbf{\begin{tabular}[c]{@{}l@{}}F1 \\ Score\end{tabular}}} \\
\multicolumn{1}{|l|}{} &  &  &  &  &  &  &  \\ \hline
\multirow{4}{*}{Glucosym} & Guideline & 8820 & 33.90\% & 0.02 & 0.32 & 0.95 & 0.73 \\ \cline{2-8} 
 & MPC & 8820 & 33.90\% & 0.02 & 0.33 & 0.95 & 0.73 \\ \cline{2-8} 
 & CAWOT & 8820 & 33.90\% & 0.01 & 0.21 & 0.96 & 0.84 \\ \cline{2-8} 
 & CAWT & 8820 & 33.90\% & \textbf{<0.01} & \textbf{\textless{}0.01} & \textbf{0.99} & \textbf{0.97} \\ \hline
\multirow{4}{*}{T1DS2013} & Guideline & 8820 & 39.30\% & 0.99 & \textbf{0.00} & 0.26 & 0.41 \\ \cline{2-8} 
 & MPC & 8820 & 39.30\% & 0.01 & <0.01 & 0.99 & 0.96 \\ \cline{2-8} 
 & CAWOT & 8820 & 39.30\% & 0.05 & <0.01 & 0.96 & 0.87 \\ \cline{2-8} 
 & CAWT & 8820 & 39.30\% & \textbf{\textless{}0.01} & 0.02 & \textbf{1.00} & \textbf{0.98} \\ \hline
\end{tabular}%






}
\vspace{-2em}
\end{table}

To sum up, by learning tight thresholds for SCS rules, CAWT monitor achieved an improvement of 12.6\%-14.9\%  in overall F1 score over the CAWOT monitor and outperformed the Guideline and MPC monitors with 32.1\% and 31.7\% increase in average F1 score on the Glucosym simulator and 141.4\% and 2.6\% on T1DS2013 simulator, along with at least 50.0\% reduction in the FPR, while keeping FNR low.




\subsubsection{Comparison with ML-based Monitors}
Table \ref{tab:mlperformance} shows the overall performance of the CAWT monitor versus three ML-based monitors in faulty scenarios (8820 simulations on each of the Glucosym and T1DS2013 simulators) using both sample level and simulation level metrics.




\begin{table}[b]
\vspace{-1.5em}
\centering
\caption{Performance of CAWT Monitor vs. ML-based Monitors}
\vspace{-0.5em}
\label{tab:mlperformance}
\resizebox{\columnwidth}{!}
{%

 

\begin{threeparttable}

\begin{tabular}{|c|l|l|l|l|l|l|l|l|l|}
\hline
\multirow{2}{*}{\textbf{\begin{tabular}[c]{@{}c@{}}Simu\\ lator\end{tabular}}} & \multicolumn{1}{c|}{\textbf{Metric}} & \multicolumn{4}{c|}{\textbf{Sample Level (Tolerance Window)}} & \multicolumn{4}{c|}{\textbf{Simulation Level (Two Regions)}} \\ \cline{2-10} 
 & \textbf{Monitor} & \textbf{FPR} & \textbf{FNR} & \textbf{ACC} & \textbf{F1 Score} & \textbf{FPR} & \textbf{FNR} & \textbf{ACC} & \textbf{F1 Score} \\ \hline
\multirow{4}{*}{\begin{tabular}[c]{@{}c@{}}Gluc\\ osym\end{tabular}} & DT & 0.08 & \textless{}0.01 & 0.93 & 0.81 & 0.56 & \textless{}0.01 & 0.57 & 0.52 \\ \cline{2-10} 
 & MLP & 0.05 & 0.03 & 0.96 & 0.86 & 0.25 & 0.02 & 0.80 & 0.70 \\ \cline{2-10} 
 & LSTM & 0.04 & 0.01 & 0.96 & 0.88 & 0.24 & 0.01 & 0.82 & 0.71 \\ \cline{2-10} 
 & CAWT & \textbf{0.01} & \textless{}0.01 & \textbf{0.99} & \textbf{0.97} & \textbf{0.12} & \textless{}0.01 & \textbf{0.91} & \textbf{0.83} \\ \hline
\multirow{4}{*}{\begin{tabular}[c]{@{}c@{}}T1DS\\ 2013\end{tabular}} & DT & 0.20 & \textbf{\textless{}0.01} & 0.83 & 0.62 & 1.00 & \textbf{<0.01} & 0.26 & 0.41 \\ \cline{2-10} 
 & MLP & 0.01 & 0.45 & 0.93 & 0.67 & 0.12 & 0.30 & 0.84 & 0.68 \\ \cline{2-10} 
 & LSTM & 0.01 & 0.03 & 0.98 & 0.94 & 0.17 & 0.03 & 0.87 & 0.78 \\ \cline{2-10} 
 & CAWT & \textbf{<0.01} & 0.02 & \textbf{1.00} & \textbf{0.98} & \textbf{0.10} & 0.01 & \textbf{0.92} & \textbf{0.87} \\ \hline
\end{tabular}%


\end{threeparttable}

 
}
\end{table}
\textbf{Sample level:}
We observe the CAWT monitor outperformed all three baseline monitors with low FPR and high accuracy and F1 score and achieved a lower FNR than the LSTM and MLP monitors on both Glucosym and T1DS2013 simulators. Although it kept a lower FNR than the CAWT monitor, the DT monitor held a much higher FPR (0.08-0.20 vs. 0.01), which will increase the risk of introducing new hazards due to unnecessary activation of the mitigation function. Overall the proposed CAWT monitor achieved the best performance among all three ML-based monitors with a 4.3\%-58.3\% increase in F1 score and 81.4\%-99.0\% reduction in FPR and retained a competitive performance, if not better, in FNR.

\textbf{Simulation level: }
Further, 
for the same number of simulations without any hazard, the DT monitor generated false alarms in 3263 (56.0\%) simulations on the Glucosym simulator and 5438 (99.7\%) simulations on the T1DS2013 simulator. In comparison, the CAWT monitor held a much lower FPR of 0.12 and 0.10 on each simulator, respectively, and thus achieved a much higher F1 score and prediction accuracy.

\subsubsection{Monitor Timeliness}

\begin{figure}[t!]
    \centering
    \includegraphics[width=0.6\columnwidth]{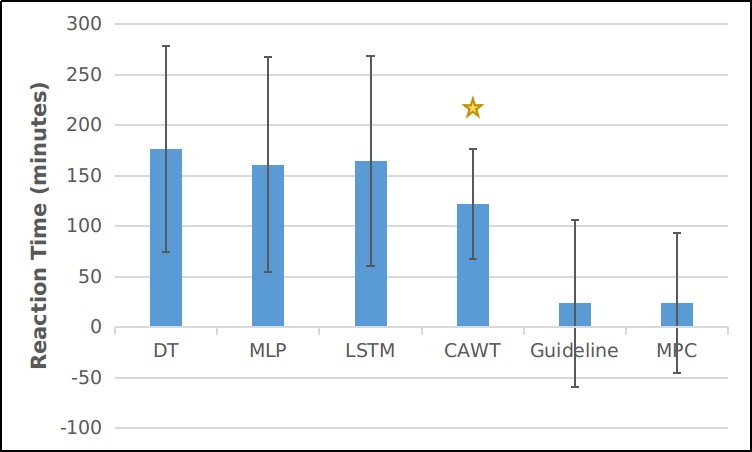}
    \vspace{-0.5em}
    \caption{Average Reaction Time for Each Monitor (minutes).} 
    \label{fig:reactiontime}
    \vspace{-1.5em}
\end{figure}

Fig. \ref{fig:reactiontime} presents the reaction time of the CAWT monitor vs. all other baseline monitors. We should emphasize that the human body is a slow system that usually takes hours to digest food and for the insulin to bring the BG value back to normal from a severe situation. Therefore, it makes sense for APS to have the reaction time measured in the order of hours (instead of seconds or minutes in other CPS with faster dynamics). We made the following  observations:
\begin{itemize}[leftmargin=*]
    \item The CAWT monitor can detect a UCA before hazard occurrence for about two hours on average, which is at least 1.6 hours longer than the MPC and Guideline monitor.
    \item The CAWT monitor kept the lowest standard deviation of reaction time, representing a more stable performance on ensuring safe reaction time for the patients. In contrast, Guideline and MPC monitors have a very high standard deviation, showing the disadvantages of not being context-aware or patient-specific. 
    \item Although their average reaction time was about 40 minutes longer than CAWT monitor's, ML-based baseline monitors' performance was not stable, and their early detection rate was 0.4\%-4.3\% less than the proposed CAWT monitor.
    
    
\end{itemize}

\subsubsection{Hazard Mitigation} 
\label{sec:mitigation-result}
We compared the mitigation performance of the CAWT monitor with the following monitors: DT monitor, which has the longest reaction time, MLP monitor with almost the same F1 score as LSTM on Glucosym simulator but with simpler logic, and MPC monitor as the best non-ML-based baseline monitor. 
We reran the simulations with each monitor and the mitigation algorithm (Algorithm \ref{alg:mitigationalgorithm}) and calculated the recovery rate, number of new hazards introduced because of FPs, and the average risk. 

\begin{table}[h]
\vspace{-0.5em}
\centering
\caption{Mitigation Performance of the CAWT Monitor and Three Baseline Monitors Using the Same Mitigation Strategy}
\vspace{-0.5em}
\label{tab:mitigation}
{%
\begin{tabular}{|l|l|l|l|l|}
\hline
\textbf{Monitor} & \textbf{CAWT} & \textbf{DT} & \textbf{MLP}& \textbf{MPC}  \\ \hline
Recovery Rate & \textbf{54.0\%} & 40.3\% & 39.0\%& 4.3\%  \\ \hline
No. New Hazard & \textbf{8} & 227 &177& 123 \\ \hline

Avg. Risk & \textbf{0.02} & 0.76 &0.68&0.22  \\ \hline

\end{tabular}%
}
\vspace{-0.5em}
\end{table}

Table \ref{tab:mitigation} shows that the CAWT monitor successfully prevented 54\% of the hazards that happened previously and only introduced eight new hazards due to false alarms, thus having the lowest average risk among the monitors.
In comparison, the MPC baseline monitor's recovery rate with the same mitigation algorithm was 4.3\%, which demonstrates the disadvantages of not being context-aware and having insufficient reaction time. 
Even though it achieved the longest average reaction time, the DT monitor only prevented 40.3\% of hazards from happening and introduced the largest number of new hazards, showing the drawback of having large FPR. A similar situation occurred to the MLP monitor, except that it got a lower average risk due to having a lower FPR than the DT monitor.

These results show that: 
(1) having a reasonable enough reaction time matters in ensuring a better recovery rate; 
(2) an appropriate balance between competitive long reaction time and low FPR is more critical in improving recovery performance than merely the longest average reaction time at any price; 
(3) the proposed CAWT monitor demonstrated the best performance in mitigating hazards, and
(4) nevertheless, only having an insulin pump limited the recovery rate from being further improved in our simulations.




\subsubsection{Resource Utilization}
We ran the simulations with different safety monitors and without a monitor a thousand times and calculated the average time overhead for each safety monitor. Results showed that the CAWT monitor has the lowest average time overhead of 252.7 us among all the safety monitors, while the time overhead of MPC, Guideline, DT, MLP, and LSTM monitors was 123.9 ms, 664.1 us, 1.3 ms, 30.7 ms, and 32.6 ms, respectively.

\vspace{-0.5em}
\section{Discussion}
\vspace{-0.5em}
Our experiments provided the following key insights:

\textbf{OpenAPS control software cannot tolerate safety-critical faults.} OpenAPS is an advanced fully-automated Control-to-Target (CTT) system \cite{blauw2016review} already equipped with some safety features, but: (1) Hazards happened even without injecting any faults to it. (2) It failed to tolerate the simulated attacks and faults. In 13.8\% of the situations where hazard happened, the BG value was less than 40 mg/dL, implying severe hypoglycemia and that the patient was unable to function \cite{SevereHypoglycemia}. (3) New hazards happened even after removing the faults.

\textbf{Patient-specific models outperform the population-based model.} 
We compared the CAWT monitor's performance with the patient-specific thresholds learned from each patient's data traces versus the population-based thresholds learned from all patients' data. For the population-based model, we learned the thresholds from the data of randomly chosen 70\% patients and tested the model on the remaining 30\% of patients' data. 
As shown in Table \ref{tab:patientspecific}, the proposed CAWT monitor with patient-specific thresholds held an advantage over a population-based CAWT monitor with at most 3.1\% and 5.3\% increase in accuracy and early detection rate (EDR), respectively. Besides, the patient-specific CAWT monitor kept the FNR low with a slightly higher FPR, therefore achieved a 24.4\% higher F1 score. These results confirm the fact that each patient has different biomedical characteristics and different tolerance levels to erroneous insulin amounts injected, and the safety monitor logic needs to be refined for each patient. 

\begin{table}[b!]
\vspace{-1.5em}
\centering
\caption{Performance of the Proposed CAWT Monitor with Either Patient-specific Threshold or Population-based Threshold}
\vspace{-0.5em}
\label{tab:patientspecific}
\resizebox{\columnwidth}{!}
{%

\begin{tabular}{|c|l|l|l|l|l|l|}
\hline
\textbf{Patient} & \textbf{Threshold} & \textbf{FPR} & \textbf{FNR} & \textbf{ACC} & \textbf{F1 Score} & \textbf{EDR} \\ \hline
\multirow{2}{*}{patientA} & Patient-specific & 0.007 & \textbf{0.00} & \textbf{0.99} & \textbf{0.94} & \textbf{99.7\%} \\ \cline{2-7} 
 & Population-based & \textbf{0.006} & 0.22 & 0.97 & 0.80 & 96.6\% \\ \hline
\multirow{2}{*}{patientH} & Patient-specific & 0.008 & \textbf{0.01} & \textbf{0.99} & \textbf{0.97} & \textbf{100.0\%} \\ \cline{2-7} 
 & Population-based & \textbf{0.007} & 0.21 & 0.97 & 0.84 & 95.0\% \\ \hline
\multirow{2}{*}{patientJ} & Patient-specific & \textbf{0.005} & \textbf{0.02} & \textbf{0.99} & \textbf{0.97} & \textbf{100.0\%} \\ \cline{2-7} 
 & Population-based & 0.007 & 0.28 & 0.96 & 0.78 & 96.4\% \\ \hline
\end{tabular}%
}
\end{table}


\textbf{Adversarial training improves safety monitor performance.} 
Using the thresholds learned from fault-free data, the proposed CAWT monitor can only detect the UCAs before the hazard happened for 88.3\% of the time and failed to generate an alert for a hazardous situation in  9\% of the simulations. 
Adversarial training and refinement of SCS formulas with the faulty data improved the CAWT monitor's performance with 11.3\% in EDR and 8.5\% in overall F1 score. 

\textbf{Weakly supervised context-aware monitor outperforms ML-based monitors.}
Our experiments showed that in most situations the CAWT monitor could achieve a better or comparable performance to the ML-based monitors that we explored in this paper. There are several other advantages that a CAWT monitor has over ML-based monitors: 

\subsubsection{Binary vs. Multi-class Classification} The ML-based monitors explored here worked as binary classifiers that can only detect if a control action is safe or unsafe. However, for successful hazard mitigation, we also need to identify the type of predicted hazard a UCA would result in. For this purpose, we retrained the ML-based monitors as multi-class classifiers with the knowledge of hazard types. Results showed that each baseline monitor's performance dropped with at least a 14.3\% increase in FNR and 0.8\%-2.3\% decrease in accuracy. In contrast, the CAWT monitor's performance stayed the same as it had the knowledge of context from SCS.

\subsubsection{Data Limitation and Corner Cases}
Fully supervised ML-based monitors tend to suffer from overfitting to the datasets they have been trained on \cite{mllimitation}. For example, we evaluated their performance on datasets collected from fault-free simulations, and results showed at least a 48.9\% drop in F1 score compared to their performance on faulty data. In comparison, the F1 score of the CAWT monitor only decreased 3.9\% because it was trained using a weakly supervised approach that only uses faulty data to tighten the SCS thresholds.

\subsubsection{Application Strategies and Resource Limitations}

To implement the proposed CAWT monitor in a real application, we need to have access to the patient profile, collect data from simulation or real-time APS operation for several days, and refine the unknown thresholds for each SCS rule offline. 
At runtime, the CAWT monitor will load the learned thresholds and work as a wrapper integrated with the APS controller with very simple logic that requires minimal resources. However, the ML-based monitors need to load the pre-trained models and utilize much more resources than the CAWT monitor.

\subsubsection{Monitor Safety and Interpretability}
Neural network classifiers are black-box systems \cite{blackbox} that, by default, do not provide transparency and explainability for their decisions. They are also vulnerable to adversarial attacks, slight perturbations, and noise in the input\cite{goodfellow2014explaining} that can lead to misclassification results. But our proposed CAWT monitor relies on a weakly supervised and transparent model, which is simpler to verify, update, and protect. 

\vspace{-1em}
\section{Threats to Validity}
\vspace{-0.5em}

This paper focuses on the safety-critical faults or attacks targeting the APS control software. Any perturbations in the sensor data will potentially affect both the controller and the safety monitor's behavior. 
However, a number of glucose sensor error models \cite{Facchinettimodel,Lyviamodel,Vettorettimodel} have been explored and successfully applied to CGM sensors (e.g., Dexcom G4/G5 \cite{Facchinettimodel2,VettorettiG5} and Medtronic Enlite sensors\cite{Lyviamodel}), which can detect the disturbance in sensor data brought by environment noise or calibration error. Further, the OpenAPS control software we used can automatically adjust the control strategy based on the sensor errors reported by CGM sensors and keep the control command safe. So our proposed monitor can learn appropriate parameters from recorded data to capture the controller's behavior. Besides, several different approaches (e.g., change detection, redundant sensors, or ML models) have been proposed to protect the APS from faults/attacks that directly comprise sensors and actuators. Those sensor checking mechanisms can be integrated with our safety monitor. 


The proposed monitor's performance heavily relies on the accuracy and completeness of the generated SCSs, which might not be easy to derive for highly complex systems. However, our method only uses a subset of state variables that can fully represent the system's dynamics. 
Besides, our proposed formal framework for generating SCSs in collaboration with domain experts can reduce the chance of manual errors. 




\vspace{-1em}
\section{Related Work}
\vspace{-0.5em}
\textbf{Run-time Monitoring and Anomaly Detection in CPS:} Recent works on run-time safety monitoring in CPS focus on control invariant methods~\cite{choi_ccs_18}, dynamic invariant detection~\cite{aliabadi_artinali_2017}, application-dependent multi-level monitoring~\cite{gautham_multilevel_2020}, unsupervised anomaly detection from streaming data~\cite{AHMAD2017134, lavin2015evaluating}, and run-time safety guards that satisfy a predefined set of safety properties~\cite{wu2017safety, wu2019shield}.


\textbf{Run-time Monitoring with STL Learning:} 
%
Several recent works \cite{bartocci_2018,deshmukh2017robust,Camacho_McIlraith_2019} have focused on approaches for monitoring, learning, and control of CPS behaviors with STL. For example, \cite{JonesAnomaly} applied STL learning and monitoring to anomaly detection in CPS and \cite{Josephine2019STL} used STL learning for characterizing T1D patient behaviors. 

Our work distinguishes from these previous works in combining the STL formalism for specification of safety context with patient-specific STL learning for the design of context-aware monitors that can predict and mitigate safety hazards.



\textbf{Safety of APS:}
Previous works \cite{bequette2014fault,blauw2016review, ramkissoon2017review, kolle2019risk, li2011hijacking} have provided a comprehensive review of safety and security issues and design requirements for APS, including the common faults, possible attacks, and their outcomes along with solutions to address them. In particular, fault-tolerant and fail-safe controllers and fault detection/diagnosis mechanisms were proposed to address glucose sensor and insulin pump faults~\cite{meneghetti2019detection}. However, most previous efforts have focused on the faults and attacks targeting the sensors and actuators, rather than the APS controller, and on the development of methods that react upon the occurrence of hypo/hyperglycemia events rather than predicting hazards for timely mitigation~\cite{kolle2019risk}.

\vspace{-1em}
\section{Conclusion}
\vspace{-0.5em}
\label{sec:conclusion}
This paper presented a formal framework for the combined model and data-driven design of context-aware safety monitors that can predict and mitigate hazards in MCPS. We developed two closed-loop APS simulation systems as case studies to evaluate the proposed method. Experimental results showed that our monitor outperforms several baseline monitors developed using medical guidelines, MPC, and ML in accurate and timely prediction of hazards and has stable performance in ensuring sufficient reaction time and mitigating hazards. Future work will focus on evaluating the applicability of this approach to a broader range of MCPS and patient scenarios.



\vspace{-1em}
\section*{Acknowledgment}
\vspace{-0.75em}
This material is based upon work supported by the National Science Foundation (NSF) under Grant No. 1748737.
\bibliographystyle{IEEEtran}
\bibliography{IEEEabrv,main}
\end{document}